\documentclass[conference]{IEEEtran}
\IEEEoverridecommandlockouts
\usepackage{amsmath,amssymb,amsfonts}
\usepackage{algorithmic}
\usepackage{graphicx}
\usepackage{textcomp}
\usepackage{todonotes}
\usepackage{lipsum}
\usepackage{biblatex}
\usepackage{subcaption}
\usepackage{verbatim}

\input{preamble}

\begin{document}

\title{Contextual Multi-Task Reinforcement Learning for Autonomous Reef Monitoring
}
\author{
\IEEEauthorblockN{
Melvin Laux\IEEEauthorrefmark{1}\IEEEauthorrefmark{2},
Yi-Ling Liu\IEEEauthorrefmark{2},
Rina Alo\IEEEauthorrefmark{1},
Sören Töpper\IEEEauthorrefmark{1},\\
Mariela De Lucas Alvarez\IEEEauthorrefmark{2},
Frank Kirchner\IEEEauthorrefmark{1}\IEEEauthorrefmark{2}, and
Rebecca Adam\IEEEauthorrefmark{2}
}

\IEEEauthorblockA{\IEEEauthorrefmark{1}
Robotics Research Group,
University of Bremen,
Bremen, Germany}
\IEEEauthorblockA{\IEEEauthorrefmark{2}
Robotics Innovation Center,
German Research Center for Artificial Intelligence,
Bremen, Germany}
Correspondence: melvin.laux@uni-bremen.de
}

\maketitle

\begin{abstract}
%Modern underwater robotics provide a promising research direction to efficiently aid the conservation and restoration of marine ecosystems.
%However, the development of controllers that are able to reliably deal with the highly non-stationary and uncertain underwater dynamics remains a major challenge. 
Although autonomous underwater vehicles promise the capability of marine ecosystem monitoring, their deployment is fundamentally limited by the difficulty of controlling vehicles under highly uncertain and non-stationary underwater dynamics.
To address these challenges, we employ a data-driven reinforcement learning approach to compensate for unknown dynamics and task variations.
Traditional single-task reinforcement learning has a tendency to overfit the training environment, thus, limit the long-term usefulness of the learnt policy. 
Hence, we propose to use a contextual multi-task reinforcement learning paradigm instead, allowing us to learn controllers that can be reused for various tasks, e.g., detecting oysters in one reef and detecting corals in another.
We evaluate whether contextual multi-task reinforcement learning can efficiently learn robust and generalisable control policies for autonomous underwater reef monitoring. 
We train a single context-dependent policy that is able to solve multiple related monitoring tasks in a simulated reef environment in HoloOcean.
In our experiments, we empirically evaluate the contextual policies regarding sample-efficiency, zero-shot generalisation to unseen tasks, and robustness to varying water currents.  
By utilising multi-task reinforcement learning, we aim to improve the training effectiveness, as well as the reusability of learnt policies to take a step towards more sustainable procedures in autonomous reef monitoring.
\end{abstract}

\section{Introduction}

\begin{figure}
\centering
    \centering
    \includegraphics[width=1.0\linewidth]{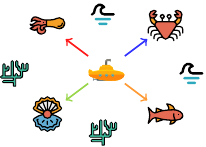}
    %\missingfigure[figwidth=0.99\linewidth]{}
    \label{fig:highlevel}
    \caption{Traditional single-task RL would require learning individual policies for monitoring different types of organisms, e.g. \(\pi_{\mathcolor{orange!80}{fish}}(a|s)\) and \(\pi_{\mathcolor{blue!80}{crab}}(a|s)\). Instead, we learn a \gradient{HSB}{unified multi-task policy}{0,240,200}{240,240,200} \(\pi(a|s;c)\) that efficiently reuses shared knowledge between tasks.}
\end{figure}

Earth's oceans have been a great source of food and resources throughout human history, however, due to \textit{the blue acceleration}, the conservation and restoration of ocean health becomes an increasingly pressing challenge \cite{Jouffray2020}.
With continually rising pollution in the oceans, the need for close and continuous health monitoring of marine ecosystems is a necessary step towards protecting biodiversity \cite{Tekman2022}.
As current restoration and protection efforts mostly rely on cost-intensive ship missions and divers, a key component for boosting the efficacy of marine restoration is the advancement of technological solutions for autonomous habitat monitoring, e.g., by using low-cost, versatile autonomous underwater vehicles (AUVs) \cite{Danovaro2025}.
A key challenge in underwater robotics is dealing with highly non-linear, time-dependent, and uncertain hydrodynamics. 
The movement behaviour of a given AUV is dependent on external factors such as currents, payload, and visibility conditions \cite{Cardenas2024}.
While some of these factors can be addressed via traditional control approaches and accurate modelling, a reliable and trustworthy controller requires a certain level of autonomy in order to deal with the remaining uncertainties that cannot be modelled or predicted accurately \cite{Christensen2022}. These challenges motivate existing work that address the need for control strategies that do not rely exclusively on model-based formulations and are able to deal with complex and changing conditions directly from interaction data.
One approach to tackle these types of uncertainties is to use data-driven methods like reinforcement learning (RL).
While single-task reinforcement learning often suffers from low sample-efficiency and a tendency to overfit, multi-task reinforcement learning (MTRL) offers a promising avenue towards learning more robust and generalisable control policies. 
By adopting a context-dependent MTRL approach for learning AUV control policies, we aim to efficiently learn policies that can function reliably under changing conditions and may be reused for different but related monitoring tasks.
Hence, in this work, we empirically investigate whether contextual MTRL can be utilised to efficiently learn generalisable and reusable AUV controllers for autonomous underwater reef monitoring.
Namely, we aim to answer the following research questions,
\begin{itemize}
    \item \textbf{RQ1}: Are RL-based policies able to solve the reef monitoring task?
    \item \textbf{RQ2}: How do contextual policies compare to mixture-of-expert policies in terms of asymptotic performance, sample-efficiency, and zero-shot generalisation to unseen tasks?
\end{itemize}

In order to answer these questions, we mathematically formalise the autonomous reef monitoring task to apply reinforcement learning methods.
We then evaluate multiple RL approaches to solve the family of tasks in a simulated environment using the high-fidelity underwater simulator HoloOcean \cite{Potokar2022}.
Additionally, we evaluate the contextual RL approach in a simplified version of the task using Minigrid \cite{Chevalier-Boisvert2023a}.
To the best of our knowledge, this is the first work to apply MTRL for autonomous reef monitoring.

\textbf{Novel Contribution:} %The
Our main contributions are %twofold, namely,

\begin{itemize}
    \item %we provide 
    a context-dependent formalisation of the autonomous task monitoring scenario for (contextual) MTRL,
    
    \item a systematic simulation-based evaluation empirically showing the contextual MTRL applicability to autonomous AUV control. %we empirically show the applicability of the contextual MTRL paradigm to autonomous AUV control through systematic evaluation in simulation.
\end{itemize}

%\todo[inline]{Rebecca: 
%Comments Introduction - Before you start the sentence "One approach to tackle..." you actually Need a few sentences (minimum one) of summarizing the SOTA or your related work.\\ 
%}
%\mariela{A sentence was added to connect to related work. However, I think that the sentences starting from 'A key challenge.. ' already references sota challenges without reciting works from the RW section.}
%Then you in a next block (or atually I would prefer subsection called "Open Challenges" or "Research Gap" address the open research gap). In another block or again preferably subsection called "research questions" formulate the research questions you address now later in this paper (but on the first occurence in experiments.) This is an application paper: The research questions we look at should be before you present the novel contributions in the introduction already (again if you like in a subsection called research questions).\\
%In novel contributions (could also be a subsection "novel contributions") you state they would be threefold. Currently, I see two bullet points.\\ 
%As a last subsection (if needed here) you could have "Notational Conventions". 
\section{Related Work}
% \melvin{add related papers here}
% \mariela{To Melvin: some suggested literature. Please check, then I can connect everything in a few paragraphs.}
% \melvin{To Mariela: thank you! Rina will also add some papers she found here, then I'll go through them and ping you}
% \melvin{To Rina: please add the papers you collected into these categories following Mariela's example}

% \textbf{On Underwater RL and AUV control}:
% \cite{Hadi2022}, \cite{Jiang2024}, 
% \cite{Arthur2024},  
% \cite{Xi2022}, \cite{Bharti2025},  
% \cite{Ma2024}, \cite{Wang2021},
% \cite{Carlucho2018}, \cite{Carlucho2018Adaptive}, 
% \cite{Xu2022}, \cite{Yuan2021},

% \textbf{Contextual/MTRL}:
% \cite{Sodhani2021}, \cite{Teh2017}
% \textbf{Zero-shot generalization}: \cite{Kirk2023}

% \textbf{Coral reef monitoring motivation}:
% \cite{Obura2019}, 
% \cite{Sultan2025}, 
% \cite{Yang2024}, 
% \cite{Pineros2024}

% coral reef monitoring
The scope of manual reef monitoring is restricted by the physical limits of human divers, who cannot survey vast or deep areas. 
Thus, AUVs need to perform these tasks with high autonomy and accuracy \cite{Pineros2024}. 
Moreover, manual data collection often lacks the consistency needed for long-term studies as human observers introduce subjective errors \cite{Obura2019}. 
To address these challenges, specialised AUVs are increasingly employed for reef monitoring, such as inspection vehicles for coral assessment or larger survey platforms that integrate multi-modal sensing with real-time AI \cite{Sultan2025}.
However, these systems are often challenged by low-visibility conditions and limited maneuverability \cite{Sultan2025}. 
Recent work has demonstrated the potential for vision-guided AUVs to map biological hotspots by correlating ecological abundance with reef topography \cite{Yang2024}. 
Scaling such approaches to unseen reef environments, however, remains a challenge.
This further highlights the need for more robust control policies that can function reliably under the unpredictable dynamics of varying underwater environments.

% underwater rl and auv control
RL has emerged as an alternative to traditional control for managing these unpredictable dynamics. 
Early work by Carlucho et al. \cite{Carlucho2018, Carlucho2018Adaptive} has demonstrated that end-to-end deep RL could handle low-level AUV control and position tracking by learning directly from raw sensor data. 
RL has been applied to specific tasks such as 3D path following \cite{Ma2024, Jiang2024}, navigation under the influence of ocean currents \cite{Wang2021, Arthur2024}, and complex path planning for the Internet of Underwater Things \cite{Xi2022}. 
Furthermore, RL-based obstacle avoidance has addressed the challenge of safe navigation by enabling AUVs to learn collision-free paths through unknown environments \cite{Yuan2021, Xu2022, Hadi2022}.
Due to the costs and risks of physical deployment, recent research has focused on using high fidelity physics simulations to develop high precision maneuvers, such as autonomous docking \cite{Bharti2025, Patil2021}.
These works demonstrate that virtual environments provide a safe and systematic way to compare RL architectures prior to their deployment. However, these simulation studies often focus on specific navigation goals, overlooking the requirement for a single controller to manage diverse objectives.

Despite these advancements, most existing RL-based controllers are limited to single-task scenarios, which often leads to overfitting and poor performance in unseen environments. 
To overcome this, Multi-Task Reinforcement Learning offers a framework for learning versatile policies that generalise across a broad range of monitoring scenarios and environmental settings. 
To improve data efficiency and stability in such settings, Teh et al. \cite{Teh2017} proposed a distilled policy approach. 
This method captures common behaviours across tasks, preventing the negative interference often seen during joint training. 
Furthermore, Sodhani et al. \cite{Sodhani2021} demonstrated that providing task metadata as context enables more efficient knowledge transfer across related tasks, which allows the agent to adjust its internal strategy to the specific requirements of a mission. While effective on standardised benchmarks, these frameworks have yet to be applied to the hydrodynamic and perceptual challenges of reef monitoring.

Learning such adaptable policies is essential for achieving zero-shot generalisation, where an agent must succeed in novel situations at deployment time without the need for on-site retraining \cite{Kirk2023}.
As noted by Kirk et al. \cite{Kirk2023}, this adaptability is required to successfully deploy RL in real-world scenarios that are complex and unpredictable. 
By employing a contextual MTRL approach, we aim to address the scaling challenges identified in recent reef studies \cite{Yang2024}, enabling a single AUV controller to function reliably across unseen sites and diverse monitoring objectives.

\section{Preliminaries}

In traditional single task reinforcement learning \cite{Sutton2018}, the agent's task is formalised as a Markov Decision Process (MDP) \(\mathcal{T}=(\mathcal{S}, \mathcal{A}, P, R, \mu)\), where \(\mathcal{S}\) is the state space, \(\mathcal{A}\) is the action space, \(P\) are the state transition probabilities, \(R\) is the reward function and \(\mu\) is the initial state distribution \cite{Bellman1957}.
The goal of an RL algorithm is then to find an optimal policy \(\pi^{*}(\mathbf{a}|\mathbf{s})\), where any policy \(\pi(\mathbf{a}|\mathbf{s})\) is a probability distribution defining the probability of selecting action \(\mathbf{a} \in \mathcal{A}\) given any known state \(\mathbf{s} \in \mathcal{S}\). Any optimal policy \(\pi^{*}\) maximises the RL objective \(J(\pi,\mathcal{T})\), i.e., the expected discounted return of policy \(\pi\) in task \(\mathcal{T}\), 
\begin{equation}
    J(\pi, \mathcal{T}) = \mathbb{E}_{r_t \sim \pi,\mathcal{T}}\left[\sum_{t=0}^{H} \gamma^{t}r_t \right], 
\end{equation}
where \(H\) is the episode horizon, \(\gamma\) is the discount factor, and \(r_t\) is the reward collected at time step \(t\) following policy \(\pi\) in task \(\mathcal{T}\).
Hence, the optimal policy is defined as 
\begin{equation}
    \pi^{*}(\mathbf{a}|\mathbf{s}) = \arg\max\limits_{\pi} J(\pi,\mathcal{T}). 
\end{equation}
In multi-task reinforcement learning, the goal is to learn a policy that is able to act optimally in a set of related MDPs \cite{VithayathilVarghese2020}.
One common assumption is to assume that the shared structure between the tasks is that all MDPs to be solved use the same state and action spaces.
Using this assumption, we can formalise the the MTRL problem as a contextual Markov decision process (CMDP) \cite{Hallak2015,Modi2018}.
A CMDP is a tuple of the form \((\mathcal{C}, \mathcal{S}, \mathcal{A}, \mathcal{M})\), where \(\mathcal{C}\) is the context space, \(\mathcal{S}\) the state space, \(\mathcal{A}\) the action space, and \(\mathcal{M}:\mathcal{C}\mapsto \mathcal{T}^{(\mathbf{c})}\) a mapping from context to a specific MDP \(\mathcal{T}^{(\mathbf{c})} = (\mathcal{S}, \mathcal{A}, P^{(\mathbf{c})}, R^{(\mathbf{c})}, \mu^{(\mathbf{c})})\), where \(P^{(\mathbf{c})}, R^{(\mathbf{c})}, \mu^{(\mathbf{c})}\) are the context-conditioned state transition probabilities, reward functions, and initial state distributions, respectively.
It should be noted that defining a suitable context representation is non-trivial for many tasks and can be arbitrarily informative ranging from simple one-hot encodings or random seeds to highly informative task domain knowledge such as physical parameters of the environment \cite{Benjamins2023}.
Assuming the context for each task is known, the goal now becomes to find an optimal context-dependent policy \(\pi^{*}(\mathbf{a}|\mathbf{s};\mathbf{c})\) that maximises the expected discounted return over all context-MDPs \(\mathcal{T}^{(\mathbf{c})}\), namely,
\begin{equation}
    \pi^{*}(\mathbf{a}|\mathbf{s};\mathbf{c}) = \arg\max\limits_{\pi} \mathbb{E}_{\mathbf{c} \sim \mathcal{C}}\left[ J(\pi, \mathcal{T}^{(\mathbf{c})}) \right].
\end{equation}
Additionally, to evaluate the zero-shot generalisation abilities of policies, it is possible to split the set of tasks induced by the the CMDP into distinct training and test sets of tasks \cite{Kirk2023}.
%\todo[inline]{
%Comments Rebecca Preliminaries - Above and in (1) and in (2): a,s not properly introduced. Do you mean A,S?\\
%c in (2) not properly introduced. I guesss you mean a is an action drawn out of A, c a cntext out of C, and s a state out of S. This all needs to be introduced nevertheless,.\\ 
%Use in ($\arg\max\limits_{\pi}$)\\
%Can you underbrace the second expectation in (2) and relate it to (1) or something explanatory?
%}
\section{Contextual MTRL for autonomous reef monitoring}
\label{sec:method}
For an AUV to be able to autonomously monitor an underwater habitat, the crucial challenge to navigate its environment to find and detect interesting regions, e.g., different types of organisms, and then navigate towards them.
We formalise this task in from of a CMDP, where the agent receives positive rewards for detecting a previously undiscovered marine organism at each time step, while receiving negative rewards for straying too far from the search area.
The task of the AUV control policy is the to select an action to maximise the likelihood of finding unseen organisms based on the current observation.

Specifically, we define a high-level representation of the state space \(\mathcal{S}\) in terms of the AUV's current location \(\mathbf{x} \in \mathbb{R}^{2}\), rotation \(\mathbf{r} \in \mathbb{R}^{2}\), and velocity \(\mathbf{v} \in \mathbb{R}^{2}\).
Additionally, the state contains the total elapsed time \(t \in \mathbb{R}_{\geq0}\) since the beginning of the episode, the number of organisms (both new and previously detected) in its immediate surrounds, as well as the percentage \(\mathbf{p} \in [0,1]^4\) of organisms of each type that still remain undiscovered.
In our setting, we consider a scenario where 500 organisms of four different types (represented by different colours, i.e., red, blue, green, black), inhabit the environment and the agent's task is to find all organisms that are flagged as interesting in the given task context.
All organisms are considered to be located on the ground. 
Hence, we omit the \(z\)-component of both location and velocity in the state space and let the AUV keep a fixed distance to the seabed.
To represent the locally detected organisms, we consider all organisms for each type that are below a given distance threshold \(d_{max}\) detected by a simulated detector oracle to focus on the challenge of navigation and control rather than perception.
The simulated detector outputs vectors containing the number of total number \(\mathbf{l} \in \mathbb{Z}_{\geq0}^{16}\) and new organisms \(\mathbf{n} \in \mathbb{Z}_{\geq0}^{16}\), in four directions (front right, front left, back right, back left).
This representation results in a 44-dimensional state space with states of the form %\(
\begin{align}
\mathbf{s} = (\mathbf{x}^{\top},\mathbf{r}^{\top},\mathbf{v}^{\top},t,\mathbf{p}^{\top},\mathbf{l}^{\top},\mathbf{n}^{\top})^{\top}.
\end{align}
%\).

In our setting, we use a generic torpedo-shaped AUV as is shown in Fig. \ref{fig:auv_closeup} with one thruster and four fins.
The action space is a discrete set of five nominal actions %\(
\begin{align}
\mathcal{A}=\{\text{forward},\text{turn right}, \text{turn left}, \text{backward}, \text{no op}\},
\end{align}
%\), 
which are converted into low-level thruster and fin commands which are then executed at a control frequency of 2 Hz.
Specifically, the forward and backward actions keep the fins in a neutral position and set the thruster to 50 or -50 percent of the maximum thrust, respectively.
The turn actions are executed by setting the vertical fins to their maximum swing in either direction, i.e., to 45 or -45 degrees, while applying a forward thrust of 50 percent.

As stated, we aim to learn policies that are able to deal with varying water currents and different types of organisms to find.
Under the assumption that currents remain constant during a single episode, we consider every different setting of currents to be a new task for the MTRL policy. 
We represent the currents for a given task as a three dimensional vector \(\mathbf{c}_{\text{c}} \in \mathbb{R}^3\).
To indicate which organism types are to be detected by the agent in any given task, we use a 4D binary vector \(\mathbf{c}_{\text{o}} \in \mathbb{B}^4\), where each entry indicates if the corresponding organism type is interesting (vector entry 1) or uninteresting (vector entry 0).
For example, the vector \((1,1,1,1)^\top\) represents the task where all types need be found and the vector \((0,1,1,0)^\top\) indicates that only blue and green organisms are of interest.
The full context space \(\mathcal{C}\) is then the Cartesian product of these two subspaces, namely, \(
\mathcal{C} = \mathbb{R}^3 \times\mathbb{B}^4\) with %\(
\begin{align}
\mathbf{c} = (\mathbf{c}_{\text{c}}^{\top}, \mathbf{c}_{\text{o}}^{\top})^{\top}\quad \forall \mathbf{c} \in \mathcal{C}.
\end{align}
%\).

Based on this context representation, we can now define the context-dependent tasks in terms of reward functions \(R^{(c)}\).
To avoid overfitting, we keep the reward function simple and define straight-forward success and and failure conditions which each lead to fixed rewards of \(r_{\text{success}}=1000\) and \(r_{\text{fail}}=-1000\).
These specific values were chosen to ensure that successful episodes always lead to positive returns and failed episodes always lead to negative returns, even in the presence of intermediate rewards.
The success condition is fulfilled once the agent has detected all organisms of interest.
The failure condition is met if the agent strays away too far from its initial position, defined by a maximum allowed distance \(d_{\text{f}}=10\).
To avoid extremely sparse rewards leading to a hard exploration problem, the agent also receives small intermediate rewards \(r_{\text{n}}=1\) whenever a new organism is detected.
In order to keep all tasks on equal scales, we normalise the intermediate rewards by the number of total organisms of interest.
Additionally, we also add a small movement regularisation term \(r_{\text{p}}=0.1\) at each time step to encourage quicker exploration strategies. Combining the intermediate reward and the regularization reward into a joint reward $r_{\text{tmp}}$
\begin{align}
r_{\text{step}}=\frac{\mathbf{n} ^\top\mathbf{c}_{\text{o}}}{\|\mathbf{c}_{\text{o}}\|_1} r_{\text{n}} - r_{\text{p}}.
\end{align}
Combining all previously discussed terms, we arrive at the context-dependent reward function,
\begin{equation}
    R^{(\mathbf{c})}(\mathbf{s}_t,\mathbf{a}_t,\mathbf{s}_{t+1})=  
    \begin{cases}
        r_{\text{success}} & \text{if } \mathbf{p}^\top \mathbf{c}_{\text{o}} = 0 \\
        r_{\text{fail}} & \text{if } \|\mathbf{x}_{t+1} - \mathbf{x}_0\|_2 \geq d_{\text{f}}\\
        %\frac{\mathbf{n} ^\top\mathbf{c}_{o}}{\|\mathbf{c}_{o}\|_1} r_n - r_p, 
        r_{\text{step}}& \text{otherwise}
    \end{cases}.
\end{equation}

Using these definitions, we now have a full description of the available context-dependent tasks for our reef monitoring scenario.
It should be noted that while the transition distribution is also dependent on the context vector by means of the water currents, it is not necessary to explicitly define them as they are assumed to be unknown in the RL setting.
Further note that due to the continuous nature of the current definition, the described CMDP induces an infinite set of tasks to sample from.

%\todo[inline]{
%Rebecca: Comments Method - All vector lengths and matrix sizes need to be introduced. Is rotation specified by a vector? Or a matrix?\\
%Defining a vector of lengths in capital bold letter is a bit weird. I mean $\mathbf{N}$. I would expect this to be a matrix when reading it, but it is a vector of lengths. Let us here find a better notation.Need to thinh about it for a while.\\
%Hi, can we change the notation for currents. I think anything related to c should be general context. Do we want to use $\boldsymbol{\gamma}$ for currents?\\
%Why $r_{success}=1000$? Say why you chose these values in the text.\\
%former (6) now (7) is too long for the column. It could help here to not show the round bracket dependency of state, action and next state. If you keep the dependency: Then are those not vectors? so use bold. Another work around of that former (6) now (7) is currently too long could be by introducing  another name for the third line in the cases equation in (7). Just changed that. Hope you do not mind.
%}
\section{Experiments}
In our experiments, we use both the simplified toy environment as well as the HoloOcean-based simulation to evaluate the suitability of contextual MTRL for the use case of autonomous reef exploration and monitoring via AUVs.
Specifically, we aim to answer the %following 
research questions:
\begin{itemize}
    \item \textbf{RQ1}: Are RL-based policies able to solve the reef monitoring task?
    \item \textbf{RQ2}: How do contextual policies compare to mixture-of-expert policies in terms of asymptotic performance, sample-efficiency, and zero-shot generalisation to unseen tasks?
\end{itemize}

To answer these research questions, we evaluate the sample-efficiency, zero-shot generalisation, and final performance of our joint contextual training regime and compare the results against training individual single-task expert policies. 
As suggested by Agarwal et al. (2021) \cite{Agarwal2021}, we report interquartile means (IQM) with bootstrapped 95\% confidence intervals (CI) of the undiscounted episode returns to analyse asymptotic performances and sample-efficiency.
We use double deep Q-networks (DDQN) \cite{Hasselt2016} as backbone algorithm, implemented using the open-source library rl-blox \cite{Laux2025}.
To train a contextual version of DDQN, denoted in the following as cDDQN, we concatenate the task context and the observed state as input to the Q-network.
All Q-networks in our experiments are represented as multilayer perceptrons (MLPs) \cite{Rosenblatt1958}.
As a baseline, we train individual policies for each separate task from the training set and then build a mixture of experts (MoE), that always selects the corresponding expert for the given task.
We then evaluate performance of each policy on both the training set and the test set of tasks.

\subsection{Minigrid Experiments}

\begin{figure*}
    \centering
    \begin{subfigure}{.33\linewidth}
        \includegraphics[width=0.99\linewidth]{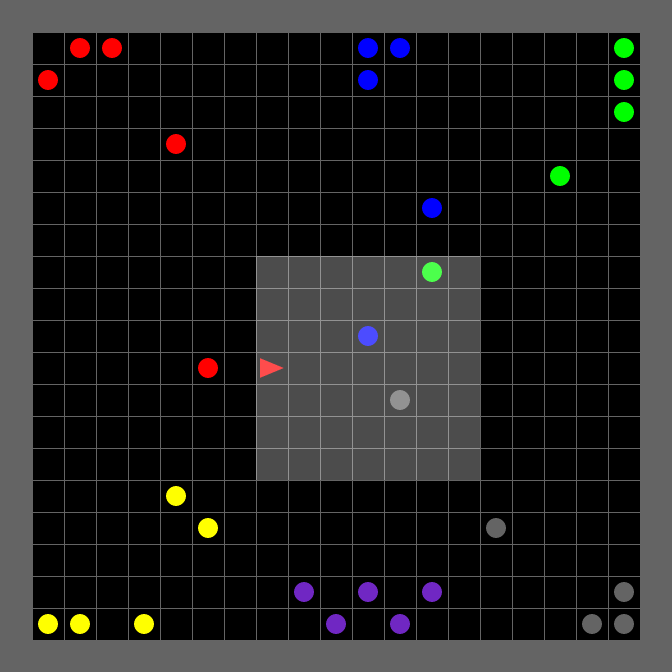}
        \caption{Fixed minigrid layout}
        \label{fig:fixed_minigrid}
    \end{subfigure}%
    \begin{subfigure}{.33\linewidth}
        \includegraphics[width=0.99\linewidth]{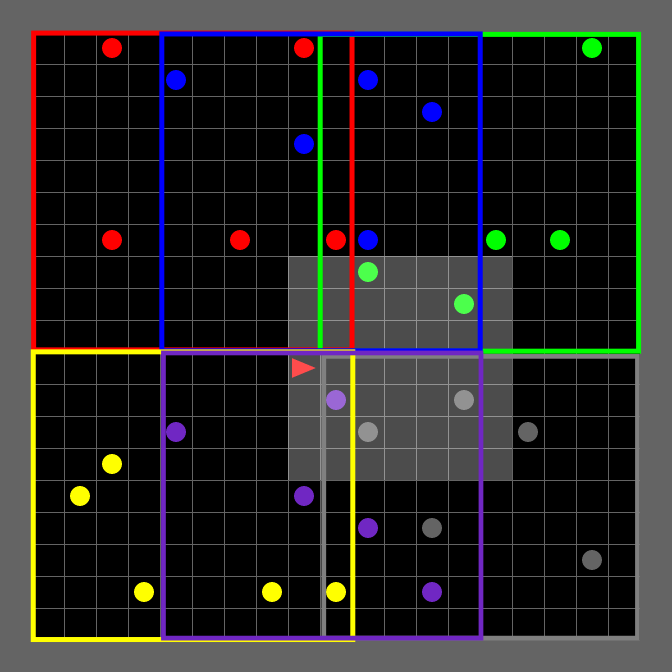}
        \caption{Organism areas for random layout}
        \label{fig:fixed_minigrid}
    \end{subfigure}%
    \begin{subfigure}{.33\linewidth}
        \includegraphics[width=0.99\linewidth]{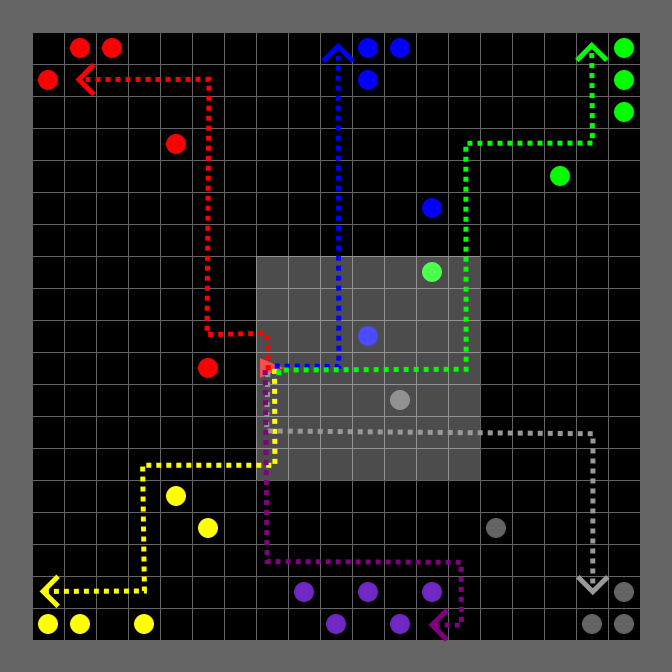}
        \caption{Examples policies}
        \label{fig:fixed_minigrid}
    \end{subfigure}

    \caption{\textbf{Minigrid environments.} For preliminary evaluation of the suitability of contextual MTRL for autonomous reef monitoring, we make use of a simplified toy setting based on minigrid. In these environments, the task of the agent is to move towards organisms (represented as circles) of the correct colour, while avoiding moving into walls or incorrectly coloured organisms. The agent receives a small positive reward for moving directly next to a correct organism, a large positive reward for finding all five correctly coloured organisms, and strong negative reward for colliding with the surrounding wall or incorrectly coloured organisms. At each timestep the agent observes its immediate surroundings as highlighted and may choose between moving forward, turning right, or turning left. (a) shows the layout of organisms in the fixed minigrid setup, (b) shows the areas in which the different coloured organisms can be placed in the random setting, and (c) shows example trajectories of final policies trained using cDDQN after 1M timesteps in the fixed setting.}
    \label{fig:minigrid_overview}
\end{figure*}

We initially test the method in a discrete approximation of the autonomous exploration task using Minigrid \cite{Chevalier-Boisvert2023a}, in which the underwater world is represented as a 2D grid world as shown in Fig. \ref{fig:minigrid_overview}.
The agent's goal is to navigate to the correctly coloured object based on the given task without running into walls or other coloured objects.
The different task contexts consist of a one-hot encoding to indicate which colour of object the agent is expected to collect, resulting in a total of six different contexts
\begin{equation}
    \mathcal{C} = \{\text{red}, \text{blue}, \text{green}, \text{yellow}, \text{purple}, \text{grey}\}.
\end{equation}
To be able to evaluate the zero-shot generalisation of the trained policies, we split the context set into a training set \(\mathcal{C}_{\text{train}}\) containing the red, green, yellow, and grey tasks and a test set \(\mathcal{C}_{\text{test}}\) containing the blue and purple tasks.
We consider two different settings of the environment, \textit{fixed} and \textit{random}. 
In the fixed setting, objects are always placed in the same locations, while in the random setting, the locations of the objects are sampled uniformly from colour-specific areas with in the grid at the beginning of each episode.
The policies in these experiments are trained for 1M timesteps in the fixed setting and for 5M timesteps in the random setting.
Each policy's Q-network is represented as an MLP with two hidden layers of 32 units each.
We repeat each experiment for both settings with 10 different random seeds.

\begin{figure*}
    \centering
    \begin{subfigure}{0.5\linewidth}
        \includegraphics[width=0.99\linewidth]{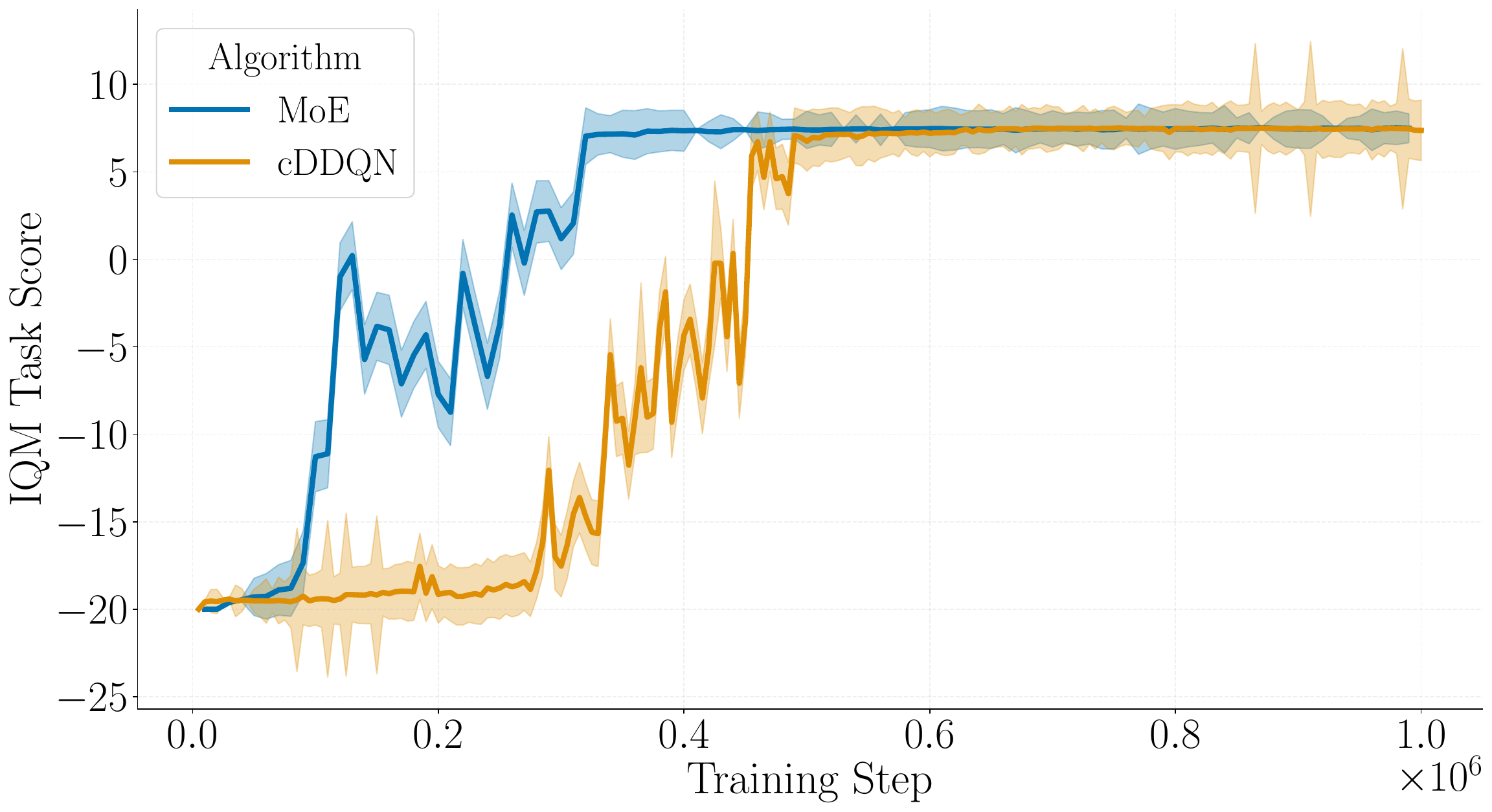}
        \caption{Training set}
        \label{fig:minigrid_fixed_train}
    \end{subfigure}%
    \begin{subfigure}{0.5\linewidth}
        \includegraphics[width=0.99\linewidth]{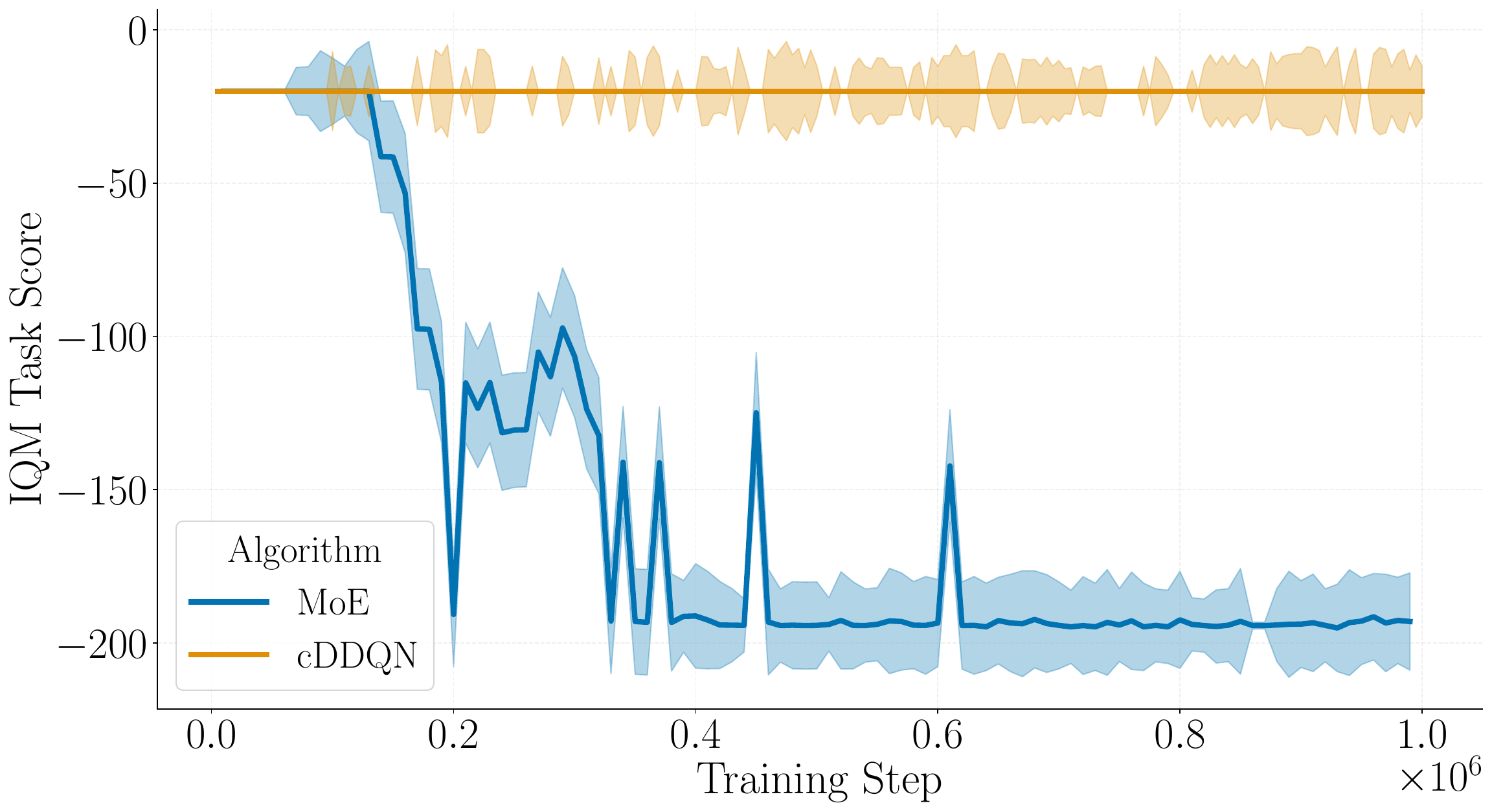}
        \caption{Test set}
        \label{fig:minigrid_fixed_test}
    \end{subfigure}
    \caption{\textbf{Results Fixed Minigrid.} Interquartile means over 10 random seeds of trained policies on the training set (a) and the test set (b) of the fixed Minigrid environments. Shaded areas show the 95\% CI of the IQM. Both MoE and cDDQN are able to solve the training tasks. However, MoE overfits and fails to avoid collisions on the unseen test set, while cDDQN is able to avoid catastrophic failure.}
    \label{fig:minigrid_fixed}
\end{figure*}

Fig. \ref{fig:minigrid_fixed} shows the evaluation performance of the trained policies on the training and test sets over time in the fixed setting.
On the training tasks, both cDDQN and MoE achieve the same asymptotic performance, however, the MoE policy is able to achieve this with fewer training steps (Fig. \ref{fig:minigrid_fixed_train}).
On the unseen test tasks, the performance of cDDQN remains stagnant at initial level, while the performance of MoE significantly drops as training progresses (Fig. \ref{fig:minigrid_fixed_test}).
We hypothesise that this effect is caused by overfitting as the cDDQN policy is aware to be in a new context, while MoE overconfidently follows the same strategies in test tasks as in training tasks leading to collisions with incorrect organisms.

\begin{figure*}
    \centering
    \begin{subfigure}{0.5\linewidth}
        \includegraphics[width=0.99\linewidth]{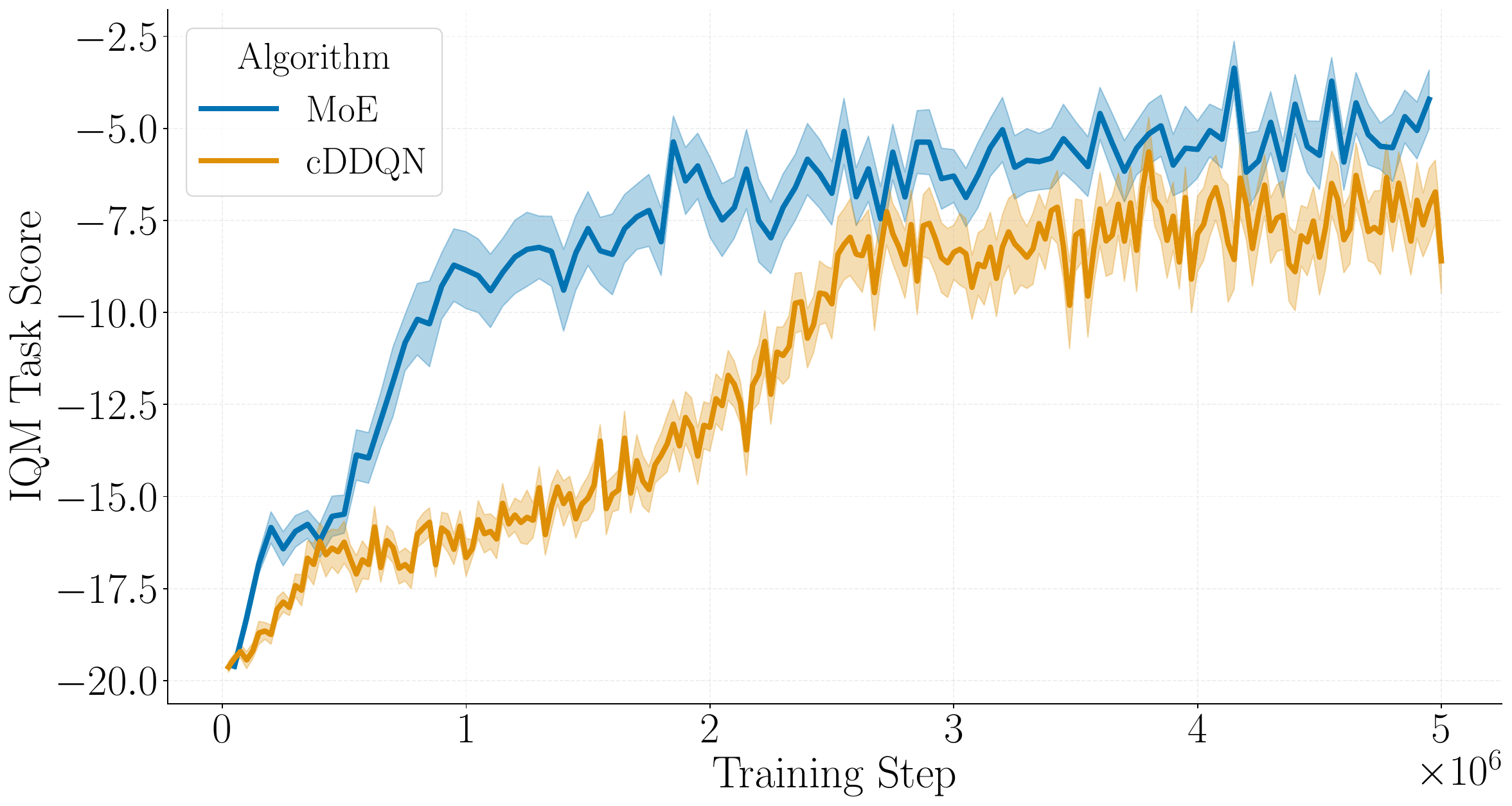}
        \caption{Training set}
        \label{fig:minigrid_random_train}
    \end{subfigure}%
    \begin{subfigure}{0.5\linewidth}
        \includegraphics[width=0.99\linewidth]{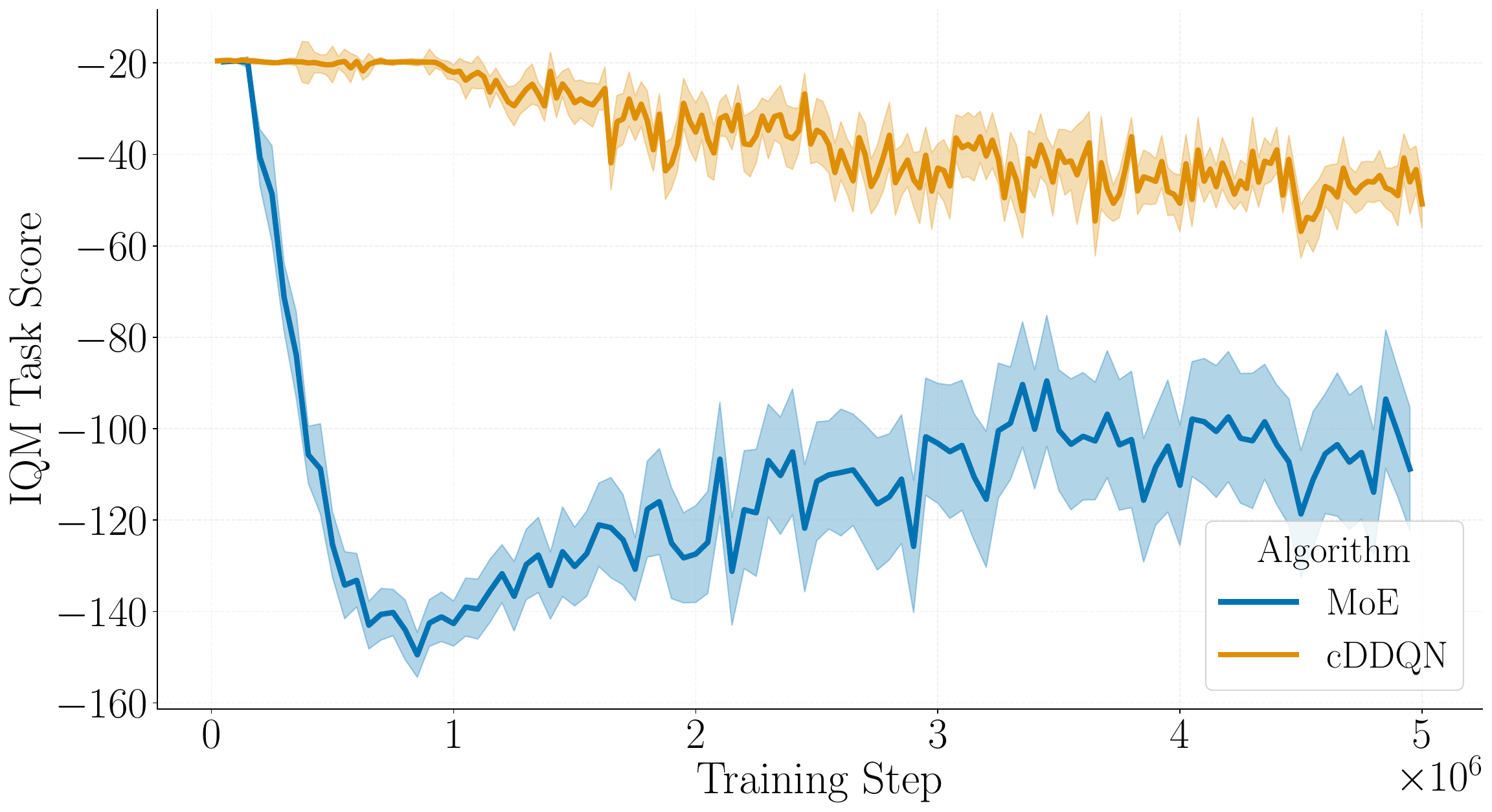}
        \caption{Test set}
        \label{fig:minigrid_random_test}
    \end{subfigure}
    \caption{\textbf{Results Random Minigrid.} Interquartile means over 10 random seeds of trained policies on the training set (a) and the test set (b) of the random Minigrid environments. Shaded areas show the 95\% CI of the IQM. Both MoE and cDDQN are able to learn policies to correctly find organisms in the training tasks with MoE slightly outperforming cDDQN both in terms of sample efficiency and asymptotic performance. However, as in the fixed setting MoE overfits and fails to avoid collisions on the unseen test set, while cDDQN is able to mostly avoid catastrophic failures.}
    \label{fig:minigrid_random}
\end{figure*}

Similarly, Fig. \ref{fig:minigrid_random} shows the performance of the trained policies on both task sets in the randomised setting. 
Each policy was evaluated on randomly generated grids, reporting the mean total reward on each task over 25 rollouts.
As in the fixed setting, both MoE and cDDQN are able to learn policies that avoid collisions and find organisms of the correct type.
However, in this more complex scenario, neither algorithm is able to perfectly find every organism of the correct type in every evaluation. 
As in the previous setting, MoE has a better asymptotic performance on the training set and a better sample efficiency than cDDQN (Fig. \ref{fig:minigrid_random_train}), but a much worse zero-shot generalisation to the unseen test tasks (Fig. \ref{fig:minigrid_random_test}).

\begin{comment}
\begin{figure}
    \centering
    \includegraphics[width=0.99\linewidth]{figures/reefshield_random_train_violin.pdf}
    \caption{Performance on final policies on each individual task}
\end{figure}

\begin{figure}
    \centering
    \includegraphics[width=0.99\linewidth]{figures/reefshield_large_train_individual_uts.pdf}
    \caption{Evaluation of contextual MTRL policy on the training tasks (red, green, yellow, purple)}
\end{figure}

\begin{figure}
    \centering
    \includegraphics[width=0.99\linewidth]{figures/reefshield_large_train_individual_st0.pdf}
    \caption{Evaluation of Task 0 expert on the training tasks (red, green, yellow, purple)}
\end{figure}
\begin{figure}
    \centering
    \includegraphics[width=0.99\linewidth]{figures/reefshield_large_train_individual_st1.pdf}
    \caption{Evaluation of Task 1 expert on the training tasks (red, green, yellow, purple)}
\end{figure}
\begin{figure}
    \centering
    \includegraphics[width=0.99\linewidth]{figures/reefshield_large_train_individual_st2.pdf}
    \caption{Evaluation of Task 2 expert on the training tasks (red, green, yellow, purple)}
\end{figure}
\begin{figure}
    \centering
    \includegraphics[width=0.99\linewidth]{figures/reefshield_large_train_individual_st3.pdf}
    \caption{Evaluation of Task 3 expert on the training tasks (red, green, yellow, purple)}
\end{figure}
\begin{figure}
    \centering
    \includegraphics[width=0.99\linewidth]{figures/reefshield_large_test_individual.pdf}
    \caption{Evaluation on the test tasks (blue and grey)}
\end{figure}
\end{comment}

\subsection{HoloOcean Experiments}
\begin{center}
\begin{figure*}
    \centering
    \begin{subfigure}{.45\linewidth}
        \includegraphics[width=0.9\linewidth]{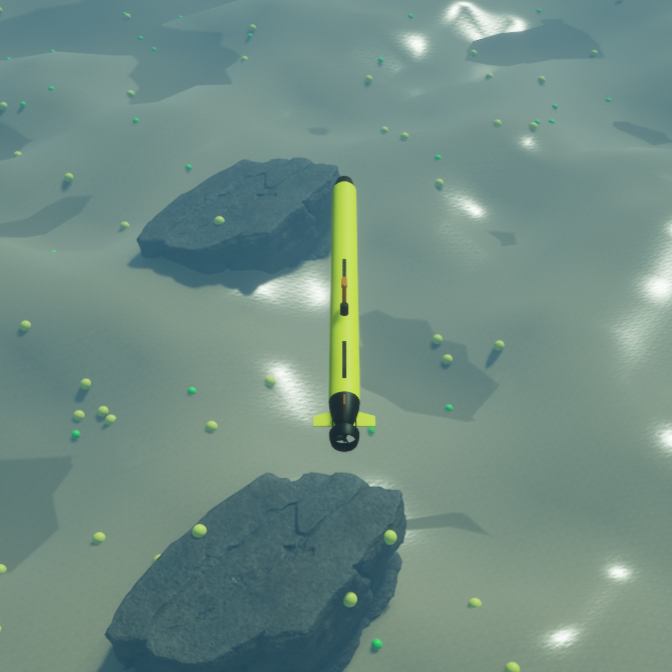}
        \caption{AUV closeup}
        \label{fig:auv_closeup}
    \end{subfigure}%
    \begin{subfigure}{.45\linewidth}
        \includegraphics[width=0.9\linewidth]{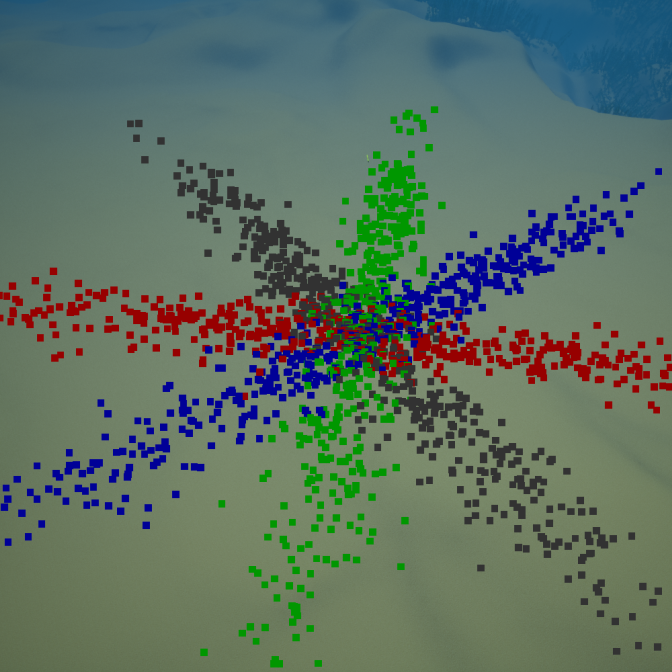}
        \caption{Organism distribution}
        \label{fig:organism_distribution}
    \end{subfigure}%

    \caption{\textbf{HoloOcean environments.} For our main investigation on contextual MTRL for autonomous reef monitoring, we implement a simulated reef environment using HoloOcean. In this environment, the task of the agent is to move towards organisms (represented as circles) of the correct colour, without leaving the search area. The agent receives a small positive reward for detecting to a correct organism, a large positive reward for finding all correctly coloured organisms, and strong negative reward for leaving the search area. A detailed formalisation of the task as a CMDP can be found in Section \ref{sec:method}. (a) shows the generic torpedo-shaped AUV used in our experiments, (b) shows an example of the organism distribution within the simulated environment.}
    \label{fig:holoocean_overview}
\end{figure*}
\end{center}
In our second set of experiments, we use a simulated environment of the reef monitoring task as introduced in the previous section using HoloOcean \cite{Potokar2022, Romrell2025}.
Figure \ref{fig:holoocean_overview} shows the used generic AUV, as well as the distribution of possible locations for the different types of organisms.
For this set of environments, we again create a training set and a test set of tasks.
The training set consists of five different current settings (north, east, south, west, and none) and a single type of organism to detect.
In combinations this leads to 20 distinct training tasks. 
For the test set, we consider four new and unseen task settings (north-east, south-east, north-west, and south-west), each combined with a single organism type.
Additionally, we also include the setting with no currents with the goal being to find organisms of any colour, i.e., the context \((0,0,1,1,1,1)\), resulting in a total of 17 test tasks.
Each policy's Q-network is represented as an MLP with two hidden layers of 128 units each.
All policies are trained for a total of 250k timesteps and repeat each experiment for 20 different random seeds.

\begin{figure*}
    \centering
    \begin{subfigure}{0.5\linewidth}
        \includegraphics[width=0.99\linewidth]{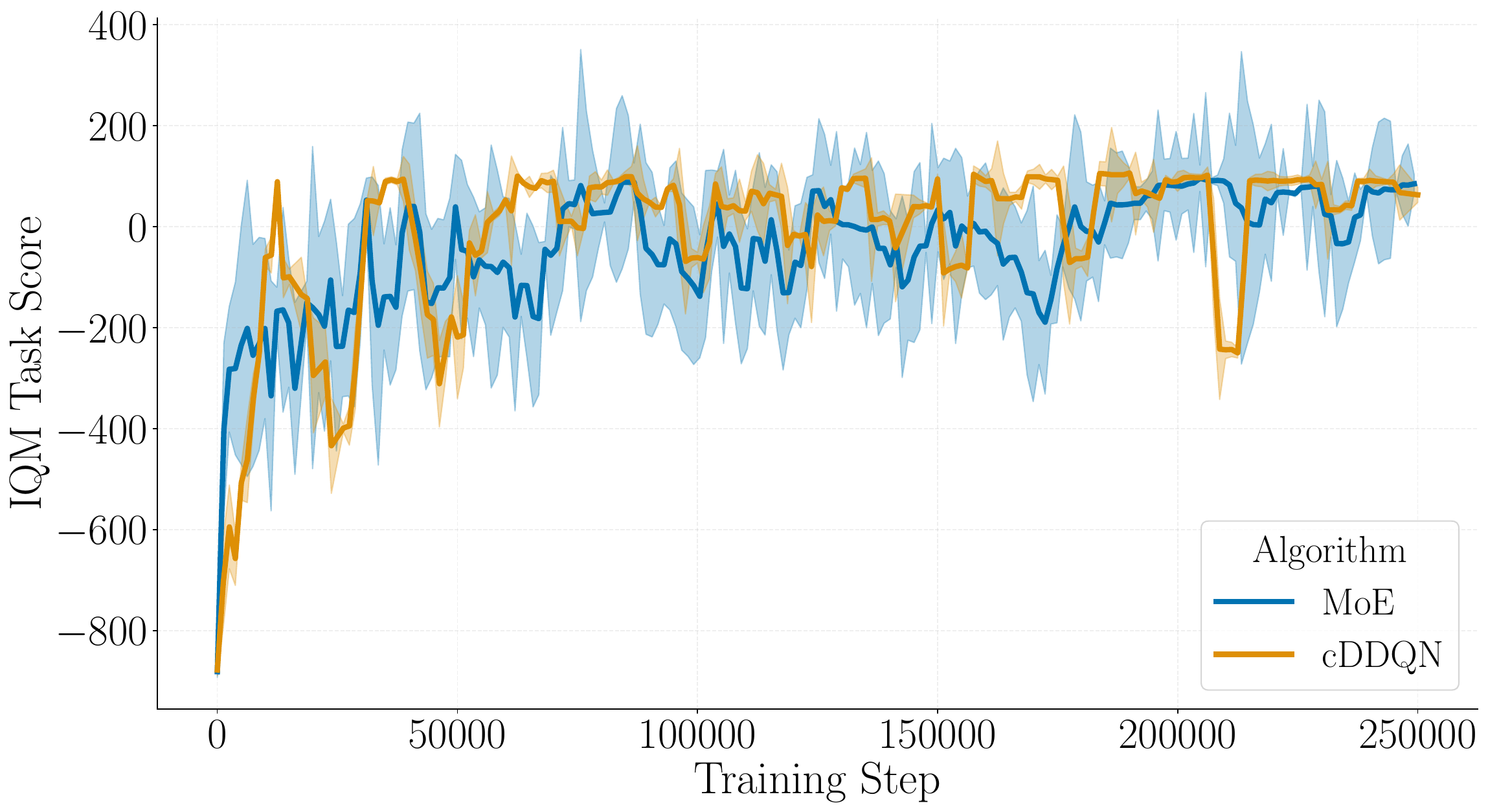}
        \caption{Training set}
        \label{fig:holoocean_train}
    \end{subfigure}%
    \begin{subfigure}{0.5\linewidth}
        \includegraphics[width=0.99\linewidth]{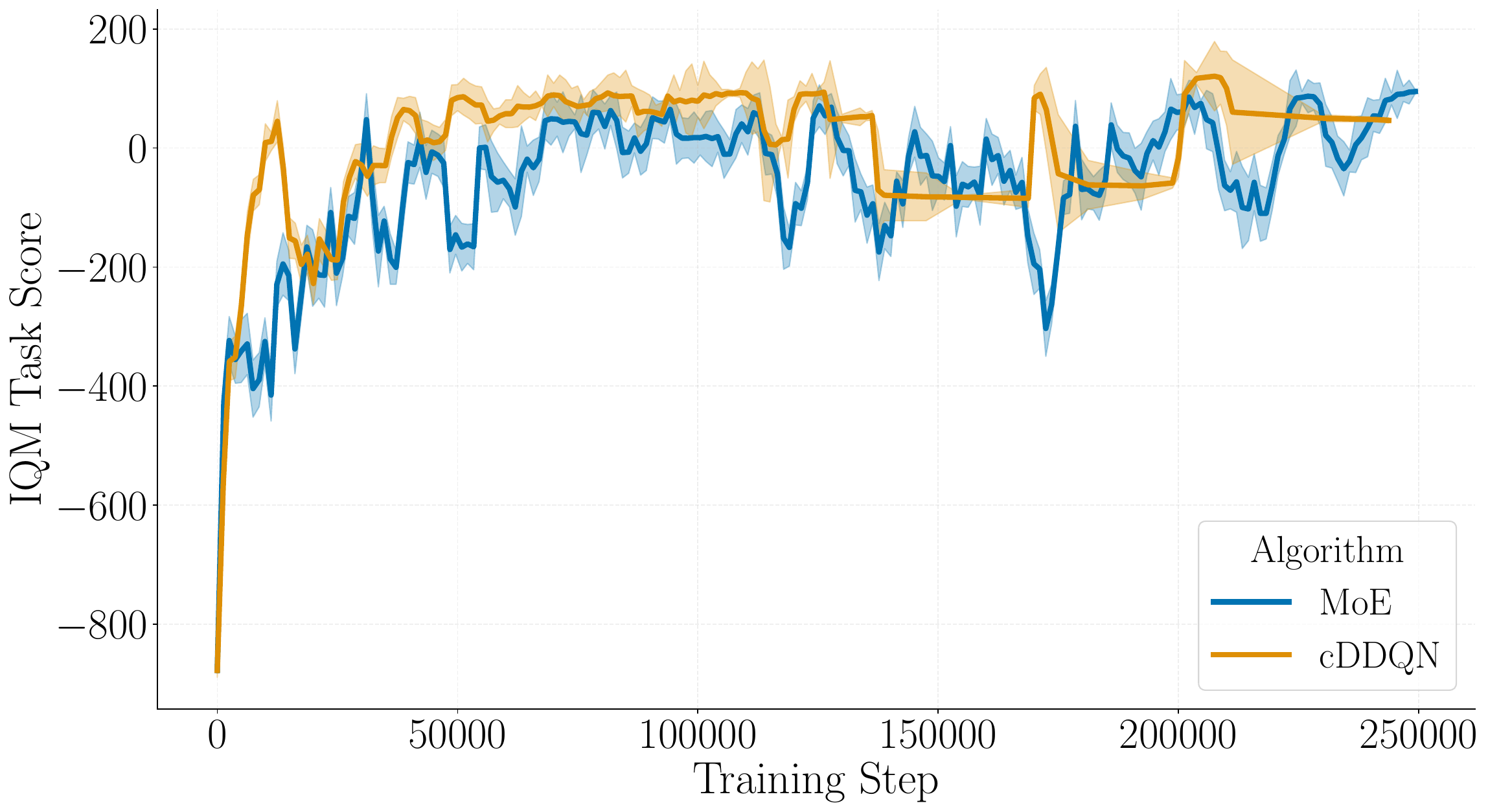}
        \caption{Test set}
        \label{fig:holoocean_test}
    \end{subfigure}
    \caption{\textbf{HoloOcean results.} Interquartile means over 20 random seeds of trained policies on the training set (a) and the test set (b) of the random HoloOcean environments. Shaded areas show the 95\% CI of the IQM. Both MoE and cDDQN are able to learn policies to correctly find organisms in the training tasks and not leaving the search area. Both approaches are able to transfer to the unseen test tasks with only a small drop in performance. This effect is likely due to the high similarity between training and test tasks.}
    \label{fig:holoocean}
\end{figure*}

Figure \ref{fig:holoocean} shows the evaluation performance of the trained policies of the HoloOcean experiments on the training and test sets over time.
Each policy was evaluated in each context for 25 rollouts of which we report the mean total reward.
Both cDDQN and MoE achieve a similar asymptotic performance on the training scenarios with no notable difference in sample-efficiency (\ref{fig:holoocean_train}). 
On the unseen test tasks, the performances of both approaches once again exhibit similar performance and sample efficiency, however on a slightly lower overall level (\ref{fig:holoocean_test}).

\subsection{Discussion}
In our extensive experiments, both in Minigrid and HoloOcean, we evaluated the performance, zero-shot generalisation, and sample efficiency of contextual MTRL compared to an MoE approach. 
Throughout our experiments, we observed similar asymptotic performance of both methods on the known training environments. 
In terms of sample-efficiency, we observed a slightly faster learning of the training tasks in the Minigrid environments for MoE.
However, the zero-shot transfer of these learnt policies to unseen training task led to catastrophic failures, i.e. collisions, likely due to overfitting, while the contextual MTRL method generated more robust policies.
In the HoloOcean scenario, we did not observe any notable differences in sample efficiency, zero-shot generalisation, or asymptotic performance.
However, it should be noted that when using the MoE approach, multiple networks are trained for every task in the training set.
In the HoloOcean experiments this means that the final MoE policies consist of 20 Q-networks, while the cDDQN policy captures all encountered tasks within a single neural network, thus, scaling down the memory requirements of the final policies by a factor of 20.
We believe this aspect to be a valuable insight, especially considering generally highly limited computational resources on a real deployed AUV.
As task complexity and sensor availability increase in future missions, resource-efficient training algorithms and model will be crucial to equip AUVs with autonomous exploration capabilities.
\section{Conclusion}

In this work, we evaluated the suitability of contextual MTRL for autonomous reef exploration and monitoring using AUVs and provided an initial mathematical formulation of a reef monitoring task in terms of a CMDP suitable for learning high-level AUV control policies using MTRL. 
In our simulated experiments, we observed promising initial results in terms of performance, sample-efficiency, and zero-shot generalisation when compared to a straight-forward mixture of experts approach.
Experiments using the HoloOcean simulator indicate that contextual policies may help improve the efficiency and robustness to unseen situations.
Notably, contextual policies may be especially promising in highly dynamic environments when limited computational resources are available. 

Rather than a ready-to-use system for immediate real-world impact, we see this work as a proof of concept for the usefulnuss of MTRL in AUV-based reef exploration.
Based on our findings, we see various potential avenues for further research.
For example, we intend to evaluate the MTRL approach more rigorously using more realistic and dynamic environments using HoloOcean, both on real systems and in field tests.
Additionally, we aim to further investigate methodological improvements from the existing MTRL literature, e.g., by using more advanced backbone algorithms and model representations \cite{Wehbe2018}, online adaptation \cite{Wehbe2019}, or exploration strategies \cite{Pathak2017,Laux2020,Amin2021}.
Finally, we also aim to investigate to close the loop between perception and control for AUV monitoring tasks to allow a joint learning of underwater object classifiers and AUV control polices.

\section*{Acknowledgments}
This work was funded by the German Federal Ministry for the Environment, Climate Action, Nature Conversation and Nuclear Safety (BMUKN) supported by the ZUG under grants 67KIA4036C and 67KIA4036A, and partially supported by the German Federal Ministry of Research, Technology and Space (BMFTR) under the Robotics Institute Germany (RIG) under grant 16ME1010.
The authors would like to thank Bilal Wehbe, Yuhan Jin, and Nayari Lessa for their valuable feedback and discussion on this manuscript. 

%\newpage
\printbibliography

\end{document}